\documentclass[10pt, a4paper]{article}

\usepackage[final]{lrec2026} 

\title{Character-aware Transformers Learn an Irregular Morphological Pattern Yet None Generalize Like Humans}

\name{Akhilesh Kakolu Ramarao\textsuperscript{1}, Kevin Tang\textsuperscript{1,3}, Dinah Baer-Henney\textsuperscript{2}}

\address{\textsuperscript{1}Department of English Language and Linguistics, Heinrich Heine University D\"{u}sseldorf \\
         \textsuperscript{2}Institut f\"{u}r Germanistik, Philologische Fakult\"{a}t, Ruhr-Universit\"{a}t Bochum \\
         \textsuperscript{3}Department of Linguistics, College of Liberal Arts and Sciences, University of Florida \\
         \{akhilesh.kakolu.ramarao, kevin.tang\}@uni-duesseldorf.de, dinah.baer-henney@rub.de\\}

\abstract{
Whether neural networks can serve as cognitive models of morphological learning remains an open question. Recent work has shown that encoder-decoder models can acquire irregular patterns, but evidence that they generalize these patterns like humans is mixed. We investigate this using the Spanish \emph{L-shaped morphome}, where only the first-person singular indicative (e.g., \textit{pongo} `I put') shares its stem with all subjunctive forms (e.g., \textit{ponga, pongas}) despite lacking apparent phonological, semantic, or syntactic motivation. We compare five encoder-decoder transformers varying along two dimensions: sequential vs. position-invariant positional encoding, and atomic vs. decomposed tag representations. Positional encoding proves decisive: position-invariant models recover the correct L-shaped paradigm clustering even when L-shaped verbs are scarce in training, whereas sequential positional encoding models only partially capture the pattern. Yet none of the models productively generalize this pattern to novel forms. Position-invariant models generalize the L-shaped stem across subjunctive cells but fail to extend it to the first-person singular indicative, producing a mood-based generalization rather than the L-shaped morphomic pattern. Humans do the opposite, generalizing preferentially to the first-person singular indicative over subjunctive forms. None of the models reproduce the human pattern, highlighting the gap between statistical pattern reproduction and morphological abstraction.\\ \newline \Keywords{Cognitive Modeling, Morphological Inflection, Morphomes, Transformers, Positional Encoding}}

\begin{document}

\maketitleabstract

\section{Introduction}

Inflectional paradigms are not merely lists of forms but constitute a central organizing principle of morphological knowledge \citep{stump2001inflectional}. This claim, that the paradigm rather than the morpheme is the primary object of morphological analysis \citep{Matthews1972}, has profound implications for how we understand the mental lexicon. If paradigms are cognitively primitive, then speakers do not assemble inflected words by concatenating stored morphemes; instead, they exploit implicational relations among whole forms, inferring unattested forms from partial paradigmatic knowledge \citep{ackerman2009parts}. Yet how far does this organization extend to irregular verbs? Can paradigmatic patterns that lack phonological, semantic, or syntactic motivation be learned and productively generalized? In this paper, we use computational modeling to probe these questions, training neural network models that differ in architectural choices on inflectional paradigms, comparing how these choices affect pattern acquisition, and evaluating their generalization behavior against human speakers tested on the same items.\footnote{Code and Data: \url{https://anonymous.4open.science/r/code_cmcl2026/}}

We focus on \emph{morphomic} patterns \citep{aronoff1994}, which provide the strongest test of paradigmatic learning because they lack any apparent phonological, semantic, or syntactic motivation. They challenge theories that require all morphological structure to be grounded in meaning or function, representing instead autonomous organization within the morphological system itself. Such patterns are well attested across Romance languages \citep{maiden2011allomorphy,maiden2018romance,maiden2021morphome}, making them ideal candidates for investigating whether computational models can acquire purely morphological generalizations.

In Spanish, the \emph{L-shaped morphome} is one such pattern: the first-person singular indicative (\textit{pongo}) shares its stem with every subjunctive form (\textit{ponga, pongas, \ldots}), while all other indicative forms use the regular stem (\textit{pones, pone, \ldots}). Table \ref{tab:l-shape-example} illustrates this pattern for the verb \textit{poner} `to put'. The defining property is the grouping of 1\textsc{sg.ind} patterns with all subjunctive forms rather than with its own mood category.

\begin{table}[ht]
\centering\small
\begin{tabular}{l l l}
\hline & \textbf{Indicative} & \textbf{Subjunctive} \\                                                                                                                                              \hline1\textsc{sg} & \textbf{pong}-o       & \textbf{pong}-a \\                                                                                                                                     2\textsc{sg} & pon-es                 & \textbf{pong}-as \\    
3\textsc{sg} & pon-e                  & \textbf{pong}-a \\
1\textsc{pl} & pon-emos               & \textbf{pong}-amos \\
2\textsc{pl} & pon-\'{e}is            & \textbf{pong}-\'{a}is \\
3\textsc{pl} & pon-en                 & \textbf{pong}-an \\
\hline                                                                                                                                            
\end{tabular}
\caption{Present tense paradigm of \textit{poner} `to put'. Hyphens mark stem-suffix boundaries. Bold marks the L-shaped stem.}\label{tab:l-shape-example}
\end{table}

Whether morphomic patterns constitute a psychological reality for speakers remains contested. \citet{Nevins2015TheRA} argue against the cognitive status of the L-shaped morphome with evidence from wug-test experiments on Spanish, Italian, and Portuguese, but subsequent work challenges this view. \citet{cappellaro2024cognitive} provide experimental evidence that Italian speakers exhibit sensitivity to L-shaped patterns in processing and production tasks. Similarly, \citet{ONeill2025} offer corpus-based evidence that gaps in Portuguese defective verbs cluster specifically in L-pattern cells rather than being randomly distributed across the paradigm. Computational modeling offers a complementary approach to this debate. \citet{ramarao-etal-2025-frequency} train neural sequence-to-sequence models on Spanish verbal inflection, providing form-morphological feature pairs as input and generating inflected forms as output. They demonstrate that these models can reproduce L-shaped stem alternations, with frequency of L-shaped verbs playing a crucial role. However, their study employed a single architecture. We extend this line of work by comparing five model architectures to ask whether design choices affect the acquisition of morphomic patterns, and then evaluating their generalization behavior against human experimental data.

We focus on two architectural dimensions that could plausibly affect the acquisition of morphomic patterns: sequential vs. position-invariant positional encoding, and atomic vs. decomposed morphosyntactic tag representations. Character-level encoder-decoder transformers have become the dominant approach for morphological inflection \citep{Wu2021ApplyingTT}, but the Unimorph schema \citep{Sylak-Glassman2016}, which provides feature specifications for inflectional morphology, treats morphosyntactic features as unordered bundles: \texttt{V;IND;PRS;1;SG} and \texttt{V;PRS;IND;SG;1} specify the same cell. This suggests that a tag should carry the same meaning regardless of its position in the sequence. We therefore compare architectures that assign tags sequential positions against ones that give all tags a fixed positional encoding, making their representation position-invariant. On the representation side, treating tags as atomic tokens requires the model to learn feature structure implicitly, whereas decomposed representations such as one-hot vectors or linguistically-informed feature geometries \citep{harley2002} encode this structure directly. To test whether these choices affect morphomic pattern acquisition, we implement five encoder-decoder transformer variants that differ along these two dimensions. Because L-shaped verbs constitute a small minority of the Spanish lexicon, we additionally vary the proportion of L-shaped verbs in training across three frequency conditions. This allows us to assess whether architectural inductive biases compensate for the natural scarcity of L-shaped verbs or only matter when such data is abundant, and whether increased exposure to L-shaped patterns during training improves generalization to novel forms. We address the following research questions:

\begin{itemize}
    \item \textbf{RQ1} How do positional encoding and morphosyntactic tag representation affect the acquisition of L-shaped morphomic patterns?
    \item \textbf{RQ2} Do models acquire the L-shaped paradigm as a structural pattern?
    \item \textbf{RQ3} Do models productively generalize the L-shaped pattern  to nonce verbs, and how does their behavior compare to that of human speakers?
\end{itemize}

\section{Background}

The SIGMORPHON shared task series has driven rapid progress in neural morphological inflection since 2016 \citep{cotterell-etal-2016-sigmorphon}, with successive iterations expanding to typologically diverse languages \citep{cotterell-etal-2017-conll, cotterell-etal-2018-conll, vylomova-etal-2020-sigmorphon}. Character-level encoder-decoder transformers have emerged as the dominant approach \citep{Wu2021ApplyingTT}, and design choices such as separate encoders for characters and morphosyntactic tags have produced better results \citep{anastasopoulos-neubig-2019-pushing, peters-martins-2019-ist}. Yet despite high reported accuracies, productive generalization remains elusive: when models are evaluated on unseen lemmas rather than forms from familiar paradigms, performance drops sharply \citep{Goldman2022UnsolvingMI, kodner2023morphological}, and predicting inflections for unseen feature combinations proves particularly challenging \citep{kodner-khalifa-2022-sigmorphon}. This gap between memorization and productive generalization suggests that current transformer-based architectures may lack the inductive biases needed to acquire the abstract representations that morphological knowledge demands.

Whether neural networks can serve as cognitive models of morphological learning has been debated since connectionist models of the English past tense \citep{rumelhart1986, kirov-cotterell-2018-recurrent}. But so far, encoder-decoder models correlate poorly with human wug-test data \citep{corkery-etal-2019-yet, mccurdy-etal-2020-inflecting}, and transformers struggle to generalize in a wug-test setting \citep{liu-hulden-2022-transformer, weissweiler-etal-2023-counting}. This gap between model and human behavior has motivated benchmarks, such as the BabyLM Challenge, which trains models on developmentally plausible datasets to probe language acquisition \citep{warstadt-etal-2023-findings, hu-etal-2024-findings}. 

The transformer architecture has no built-in sense of order, i.e., without positional encodings, it cannot, for example, distinguish \textit{pongo} from \textit{gopon} \citep{dufter2022}. The original transformer introduced sinusoidal positional encodings \citep{DBLP:journals/corr/VaswaniSPUJGKP17}. Since then, numerous alternatives have been proposed, including relative positional encoding \citep{shaw-etal-2018-self}, learned absolute embeddings \citep{devlin-etal-2019-bert}, and rotary positional encoding \citep{su2024}.

However, the necessity of positional encoding varies across tasks: \citet{wang-chen-2020-position} show that BERT without positional encoding functions essentially as a bag-of-words model in understanding tasks (classification, QA), while \citet{haviv-etal-2022-transformer} find that causal language models remain competitive even without explicit positional encoding in generation tasks, suggesting some tasks are less sensitive to position.

How morphosyntactic tags are represented has received comparatively less attention. Since \citet{kann-schutze-2016-single}, most systems concatenate morphological features as individual tokens to the character sequence and encode both through a single encoder. Other approaches separate the two inputs: \citet{aharoni-goldberg-2017-morphological} concatenate learned feature embeddings into a single vector that conditions the decoder at every step, \citet{peters-martins-2019-ist} learn separate attention distributions for lemma characters and tags, and \citet{anastasopoulos-neubig-2019-pushing} use separate encoders for lemma characters and tags. However, decomposing feature bundles into structured vectors, such as one-hot encodings or binary features from feature geometry \citep{harley2002}, remains unexplored in the literature.

\section{Methodology}

\subsection{Data}

\subsubsection{Real verbs}

We use Spanish verbal paradigms in IPA transcription from \citet{ramarao-etal-2025-frequency}. Each verb has a 12-cell present-tense paradigm crossing three persons (1, 2, 3), two numbers (singular, plural), and two moods (indicative, subjunctive). Verbs are classified as \textit{L-shaped} if the 1\textsc{sg.ind} stem matches all subjunctive stems but differs from the remaining indicative stems, and as \textit{NL-shaped} otherwise. The full dataset contains 299 L-shaped and 4,859 NL-shaped verbs.

To investigate the effect of input frequency on learning, we create three training conditions by varying the proportion of L-shaped verbs: 10\%L-90\%NL, 50\%L-50\%NL, and 90\%L-10\%NL. The 10\%L condition most closely approximates the natural distribution of L-shaped verbs in the Spanish lexicon, where they constitute roughly 7\% of verb types. For each condition, we sample 332 lemmas from the full pool and split them into training (70\%, 232 lemmas), development (10\%, 34 lemmas), and test (20\%, 66 lemmas) sets. Following \citet{Goldman2022UnsolvingMI} and \citet{kodner2023morphological}, we ensure that there is no lemma overlap between the training, development, and test sets, so that models must generalize to unseen lemmas. The L-shaped to NL-shaped ratio is maintained across all three splits.

Each training example consists of two source form-tag pairs and one target tag (see Section \ref{subsec:task}). From a 12-cell paradigm, we generate all 660 unique (unordered source pair, target) triples per lemma and subsample 25\% per training seed. We repeat the procedure with three lemma splits and four data subsamples, yielding 12 independently trained models per condition per architecture.

\subsubsection{Nonce verbs}

To compare model behavior with human generalization, we use behavioral data from the wug test experiment by \citet{Nevins2015TheRA} on 107 native Spanish speakers. In a paradigm completion task, participants were presented two forms of 15 nonce verbs exhibiting novel stem-final consonant alternations between fricatives and stops (e.g., \textit{bus-} vs.\ \textit{but-}, \textit{mif-} vs.\ \textit{mip-}) and asked to produce a third form.

Participant responses were converted from Spanish orthography to IPA transcription, and only responses matching one of the expected paradigm cells were retained. For each nonce verb, we evaluate model predictions on the same three paradigm cells tested in the human study (1\textsc{sg.ind}, 2\textsc{sg.sbjv}, and 3\textsc{sg.sbjv}), resulting in 120 test items per model run.

\subsection{Task}\label{subsec:task}

To mirror the above wug test, where participants are given two forms and must produce a third, we frame the task as two-source morphological re-inflection \citep{Kann_etal_2017_EACL}: given two source form-tag pairs from a verb's paradigm and a target feature bundle, the model must produce the corresponding target form. For example, given (\textit{pongo}, \texttt{V;IND;PRS;1;SG}) and (\textit{ponga}, \texttt{V;SBJV;PRS;1;SG}) with target features \texttt{V;SBJV;PRS;2;SG}, the model should produce \textit{pongas}. Following \citet{Wu2021ApplyingTT}, we model this as character-level sequence-to-sequence transduction. During training, models observe all source-target combinations for each lemma; at test time, they must generalize to entirely unseen lemmas.

\subsection{Model architectures}

All five models share a common encoder-decoder transformer backbone with 4 layers, 4 attention heads, embedding dimension $d = 256$, and feedforward dimension 1024. The models differ in how the encoder handles morphosyntactic tags along two dimensions: (i) whether tags receive sequential positional encoding or are assigned a fixed position, and (ii) how tags are represented. We organize the models into two classes based on positional encoding: the first class assigns tags sequential positions as it is done with characters, while the second treats tags as position-invariant. The three architectures in the position-invariant class further differ in tag representation.

\paragraph{Training.} We use the same hyperparameter settings as \citet{ramarao-etal-2025-frequency}. All models are optimized using Adam \citep{Kingma2015AdamAM} with a learning rate of 0.001 and a linear warmup schedule over 4,000 steps. We use a batch size of 400, dropout of 0.3, and label smoothing of 0.1. Training proceeds for a maximum of 10,000 update steps. At inference, we use beam search with a beam size of 5.
For each of the three frequency conditions (10\%L, 50\%L, 90\%L), we train 12 models per architecture. 

\subsubsection{Class 1: Uniform positional encoding}

\paragraph{Vanilla.} The baseline concatenates form characters and feature tags into a single flat sequence. Characters are tokenized individually (e.g., \textit{pongo} $\rightarrow$ \texttt{p o n g o}), while each morphosyntactic feature bundle is a single vocabulary entry (e.g., \texttt{V;IND;PRS;1;SG}). Both characters and tags are embedded through a shared learned embedding matrix. Both token types receive sequential sinusoidal positional encoding based on their position in the concatenated sequence. The model has no mechanism to distinguish tag tokens from character tokens \citep{ramarao-etal-2025-frequency}. 

\paragraph{Character-separated.} Identical to \textsc{Vanilla}, except that feature tags are decomposed into individual characters separated by whitespace (e.g., \texttt{V;IND;PRS;1;SG} $\rightarrow$ \texttt{V  IND  PRS  1 SG}). All tokens share the same embedding table and receive uniform sequential positional encoding.

\subsubsection{Class 2: Position-invariant tags}

The UniMorph schema \citep{Sylak-Glassman2016} treats morphosyntactic features as unordered bundles, suggesting that sequential positional encoding may be inappropriate for morphosyntactic specifications. The following three architectures assign tags a fixed positional index of 0 while characters retain sequential positional encoding. They differ in how they represent tag content.

\paragraph{Feature-invariant.} Following \citet{Wu2021ApplyingTT}, tags are embedded through the same shared matrix as in \textsc{Vanilla} but are assigned a fixed positional index of 0, while characters retain sequential positional encoding.

\paragraph{Feature-onehot.} Tags are assigned a fixed positional index of 0 and represented not as atomic tokens but as structured feature vectors. Each tag is parsed into its component features and encoded as a one-hot vector over feature categories: mood (\textsc{ind}, \textsc{sbjv}), person (1, 2, 3), and number (\textsc{sg}, \textsc{pl}), yielding a sparse 7-dimensional binary vector. A learned linear projection maps this vector to the same $d$-dimensional embedding space used by character tokens. At each position, the model selects between the character embedding and the feature projection based on token type.

\paragraph{Feature-geometric.} Like \textsc{Feature-onehot}, tags are assigned a fixed positional index of 0 and represented as structured vectors. However, instead of one-hot category vectors, tags are encoded as binary feature vectors following the feature-geometric analysis of \citet{harley2002}: person is decomposed into [$\pm$participant, $\pm$author] (1st = [+,+], 2nd = [+,$-$], 3rd =[$-$,$-$]), number into [$\pm$plural], and mood into [$\pm$indicative], yielding a 4-dimensional binary vector. This vector is projected to the embedding space through a learned linear transformation.

\subsection{Evaluation}

\subsubsection{Sequence accuracy}

We evaluate overall sequence accuracy as the percentage of predictions that exactly match the target form, and additionally report it separately for L-shaped and NL-shaped verbs.

\subsubsection{Stem accuracy by verb type} 

The distinction between L-shaped and NL-shaped verbs is defined by the stem-final consonant alternations: L-shaped verbs use one stem in the 1\textsc{sg.ind} and all subjunctive cells and a different stem elsewhere. Whether a model has learned the alternation pattern is therefore most directly assessed at the stem level. We extract stems by stripping the conjugation suffixes (for \textit{-ar}, \textit{-er}, and \textit{-ir} verb classes) from both predicted and target forms. A prediction is stem-correct if the extracted stem matches the target stem. We report stem accuracy separately for L-shaped and NL-shaped verbs.

\subsubsection{Paradigm shape analysis}\label{subsubsec:paradigm-shape}

Beyond aggregate accuracy, we ask whether models treat the 12 paradigm cells in a way that reflects the L-shaped pattern, i.e., whether they achieve \emph{structural generalization} \citep{hupkes2023taxonomy}. If a model has learned the L-shape, its stem predictions should systematically differ between L-cells (1\textsc{sg.ind} + all subjunctive) and the remaining indicative cells. We first report raw mean stem accuracy per paradigm cell across all test lemmas to examine whether models produce different stems across cells.

To quantify whether this per-cell pattern reflects the L-shape, we apply a scoring transformation to stem accuracies aggregated across all test lemmas, regardless of verb type: L-cell accuracies are kept as is, while the remaining indicative cell accuracies are inverted ($1 - accuracy$), since for L-shaped verbs an incorrect stem in a non-L indicative cell corresponds to the model producing the L-shaped stem where the regular stem is expected. After this transformation, a high score indicates that the model produced the L-shaped stem for that cell. If L-cells score high while non-L cells score low, this suggests L-shaped behavior. We then apply $k$-means clustering ($k=2$) to group the 12 cells based on these transformed scores. A model that has acquired the L-shaped paradigm should produce two clusters: one containing 1\textsc{sg.ind} and all subjunctive cells, and another containing the five remaining indicative cells.

\subsubsection{Nonce verb evaluation}

To compare model generalization with human behavior, we evaluate all models on the nonce verb stimuli from the wug test experiment of \citet{Nevins2015TheRA}. For each of the 15 nonce verbs, the model is tested on the same three paradigm cells as in the human study (1\textsc{sg.ind}, 2\textsc{sg.sbjv}, 3\textsc{sg.sbjv}), with both possible source orderings, yielding 120 test items per model run. We report stem accuracy and per-cell paradigm shape for models alongside the human behavioral data.

\section{Results}

\subsection{Real verbs}
\subsubsection{Sequence accuracy}

\begin{figure}[ht]
\centering
  \includegraphics[width=\columnwidth]{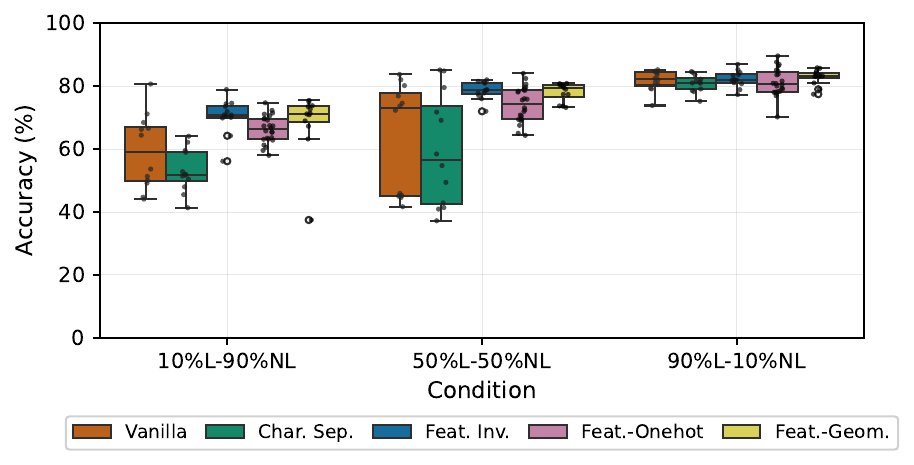}
    \caption{Overall sequence accuracy for all five architectures across the three frequency conditions.
    \label{fig:overall-accuracies}}
\end{figure}

Figure \ref{fig:overall-accuracies} compares sequence-level accuracy across the five architectures and three frequency conditions. The three position-invariant models (Class 2: \textsc{Feature-invariant}, \textsc{Feature-onehot}, \textsc{Feature-geometric}) consistently outperform the two sequential-PE models (Class 1: \textsc{Vanilla}, \textsc{Character-separated}). In the 10\%L condition, where L-shaped examples are scarce, the gap is largest: Class 2 models achieve mean accuracies of 68-70\%, whereas \textsc{Vanilla} reaches 59\% and \textsc{Character-separated} only 53\%. As L-shaped training data increases, overall accuracy improves for all architectures and the gap narrows: in the 90\%L condition, all five models converge to the 81-83\% range. Class 1 models also exhibit higher variance in the 10\%L and 50\%L conditions, whereas Class 2 models maintain low variability even when L-shaped examples are scarce. This indicates that position-invariant tag encoding provides a stronger and more stable inductive bias when L-shaped examples are scarce, but with sufficient data even simpler architectures achieve comparable performance.

\begin{figure}[ht]
\centering
  \includegraphics[width=\columnwidth]{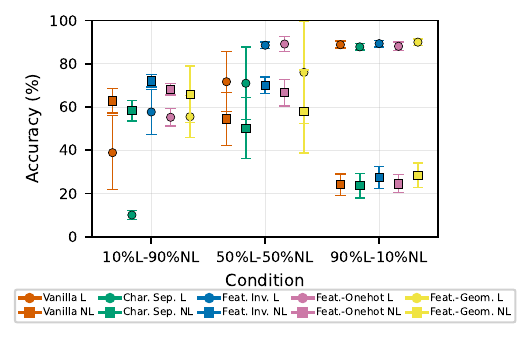}
    \caption{
    L-shape and NL-shape accuracy across five model architectures. Circles denote L-shape accuracy, squares denote NL-shape accuracy.
    \label{fig: l_nl_accuracies}}
\end{figure}

Furthermore, we examine how accurately models predict L-shaped forms compared to NL-shaped forms across the three test conditions in Figure \ref{fig: l_nl_accuracies}. As L-shaped training frequency increases, L-shaped accuracy rises, reaching 88-90\% in the 90\%L condition across all models. Conversely, NL-shaped accuracy drops as NL-shaped verbs become scarcer, falling to 24-29\% in the 90\%L condition. This symmetric pattern indicates that all models, regardless of architecture, are more strongly driven by training frequency than by the regularity of the pattern: forms from the minority verb class are consistently predicted poorly, even when that class is the regular one.

In the critical 10\%L condition, NL-shaped accuracy exceeds L-shaped accuracy for all models, reflecting the dominance of NL-shaped verbs in training. The gap is most extreme for \textsc{Character-separated} (L: 10\%, NL: 58\%) and \textsc{Vanilla} (L: 37\%, NL: 62\%), whereas Class~2 models maintain more balanced performance (\textsc{Feature-invariant}: L: 58\%, NL: 72\%; \textsc{Feature-onehot}: L: 55\%, NL: 68\%; \textsc{Feature-geometric}: L: 56\%, NL: 67\%).

\subsubsection{Stem accuracy by verb type}

\begin{figure}[ht]
\centering
  \includegraphics[width=\columnwidth]{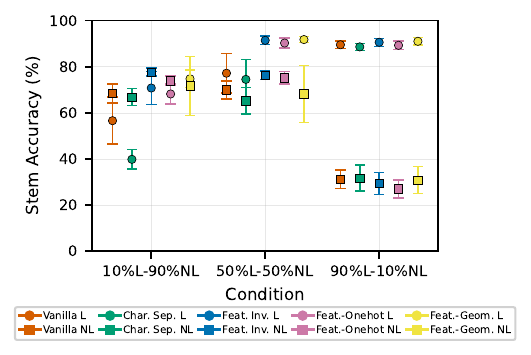}
    \caption{
    Stem accuracy separated by verb type. Circles denote L-shaped stem accuracy; squares denote NL-shaped stem accuracy.
    \label{fig:stem-accuracies}}
\end{figure}

Figure \ref{fig:stem-accuracies} evaluates stem accuracy, isolating the stem-final consonant alternation that defines the L-shaped pattern from suffix errors. Stem accuracy is consistently higher than sequence accuracy.

All models reach 89-92\% L-shaped stem accuracy in the 90\%L condition. NL-shaped stem accuracy decreases with decreasing NL training frequency, falling to 27-31\% in the 90\%L condition, mirroring the frequency dependence observed at the sequence level.

In the critical 10\%L condition, Class 2 models show a clear advantage in L-shaped stem accuracy: \textsc{Feature-geometric} (75\%), \textsc{Feature-invariant} (71\%), and \textsc{Feature-onehot} (68\%) all outperform \textsc{Vanilla} (57\%) and \textsc{Character-separated} (40\%). This indicates that position-invariant tag encoding helps models learn the correct stem alternation even from limited L-shaped exposure.

\subsubsection{Paradigm shape analysis}

\begin{figure}[ht]
    \centering
    \includegraphics[width=\linewidth]{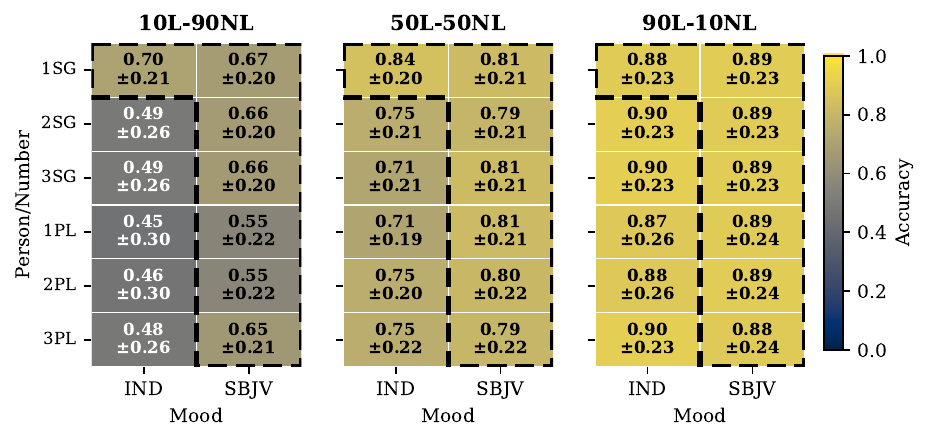}
    \caption{\textsc{Vanilla}: per-cell stem accuracy across the three frequency conditions.}
    \label{fig:paradigm-vanilla}
\end{figure}

\begin{figure}[ht]
    \centering
    \includegraphics[width=\linewidth]{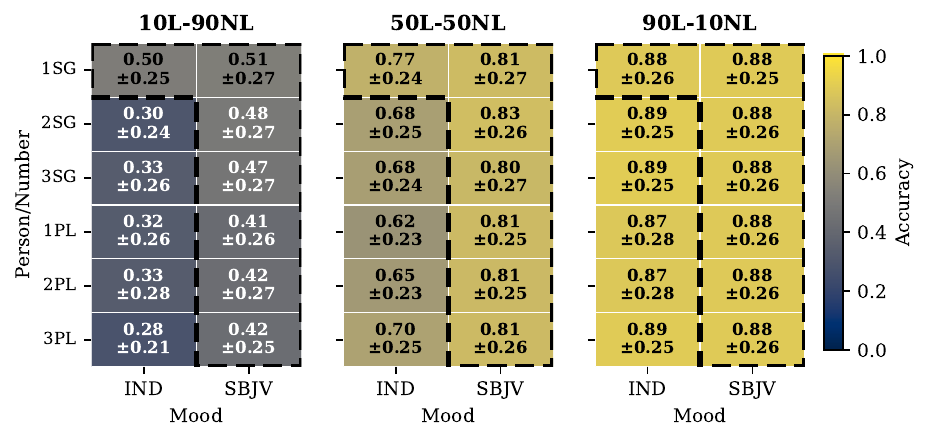}
    \caption{\textsc{Character-separated}: per-cell stem accuracy across the three frequency conditions.}
    \label{fig:paradigm-character-sep}
\end{figure}

\begin{figure}[ht]
    \centering
    \includegraphics[width=\linewidth]{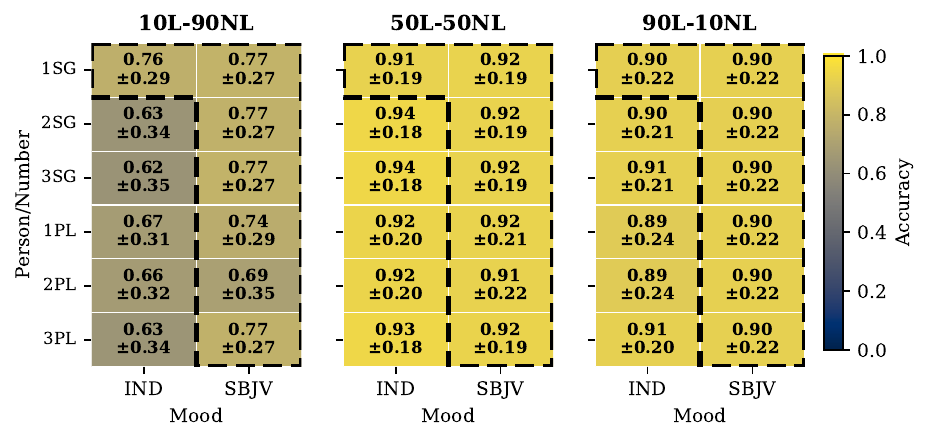}
    \caption{\textsc{Feature-invariant}: per-cell stem accuracy across the three frequency conditions.}
    \label{fig:paradigm-feature-invariant}
\end{figure}

\begin{figure}[ht]
    \centering
    \includegraphics[width=\linewidth]{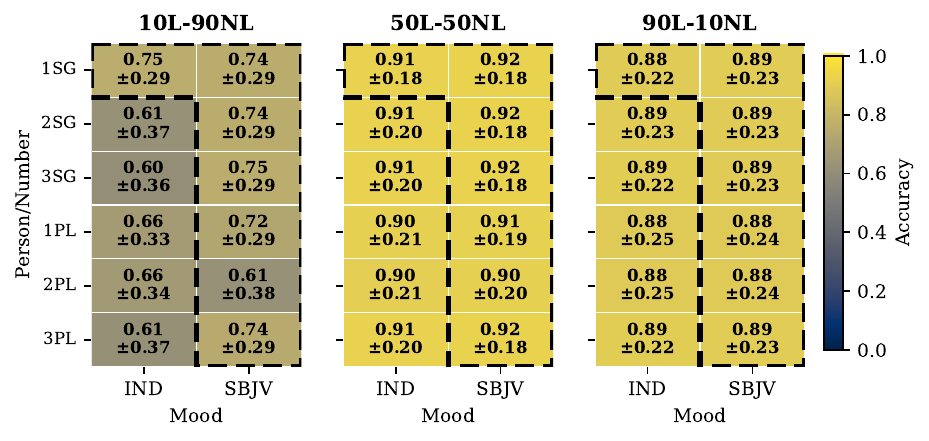}
    \caption{\textsc{Feature-onehot}: per-cell stem accuracy across the three frequency conditions.}
    \label{fig:paradigm-feature-onehot}
\end{figure}

\begin{figure}[ht]
    \centering
    \includegraphics[width=\linewidth]{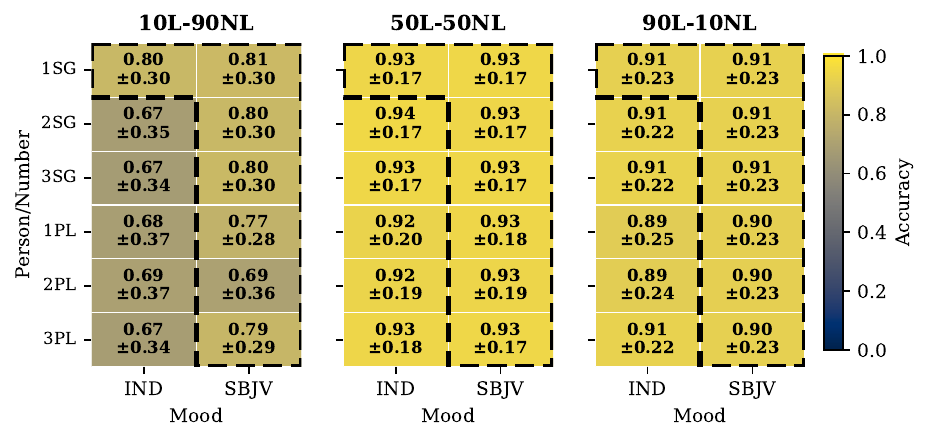}
    \caption{\textsc{Feature-geometric}: per-cell stem accuracy across the three frequency conditions.}
    \label{fig:paradigm-feature-geometric}
\end{figure}

We apply the paradigm shape analysis described in Section \ref{subsubsec:paradigm-shape}. Figures~\ref{fig:paradigm-vanilla}-\ref{fig:paradigm-feature-geometric} show raw (untransformed) per-cell mean stem accuracy ($\pm$ SD) across all test lemmas for the three conditions. Black-dashed borders indicate which cells belong to the L-shape for reference. Class 2 models show a clear accuracy contrast between L-cells and the remaining indicative cells even at 10\%L, whereas Class 1 models show more uniform accuracy across cells.

To quantify whether this pattern reflects the L-shape, we apply the scoring transformation and $k$-means clustering described in Section \ref{subsubsec:paradigm-shape}. We report results across all three frequency conditions because the critical question is not merely \emph{whether} models can recover the L-shaped clustering, but \emph{how much L-shaped exposure they require to do so}. A perfect L-shape would yield high transformed scores for L-cells (1\textsc{sg.ind} + all subjunctive) and low scores for the remaining indicative cells, producing two distinct clusters. In the 50\%L and 90\%L conditions, all five architectures recover the correct L-shaped clustering. Table \ref{tab:kmeans-10L} shows the cluster assignments in the critical 10\%L condition, where L-shaped verbs are rare in training.

\begin{table}[ht]
\centering
\resizebox{0.7\columnwidth}{!}{%
\begin{tabular}{l c c c c c c}
\hline
\textbf{Cell} & \textbf{Exp.} & \textbf{Van} & \textbf{C-Sep} & \textbf{F-Inv} & \textbf{F-1H} & \textbf{F-Geo} \\
\hline
1\textsc{sg.ind}  & L  & L  & L  & L  & L  & L  \\
2\textsc{sg.ind}  & NL & NL & NL & NL & NL & NL \\
3\textsc{sg.ind}  & NL & NL & NL & NL & NL & NL \\
1\textsc{pl.ind}  & NL & NL & NL & NL & NL & NL \\
2\textsc{pl.ind}  & NL & NL & NL & NL & NL & NL \\
3\textsc{pl.ind}  & NL & NL & NL & NL & NL & NL \\
1\textsc{sg.sbjv} & L  & L  & L  & L  & L  & L  \\
2\textsc{sg.sbjv} & L  & L  & L  & L  & L  & L  \\
3\textsc{sg.sbjv} & L  & L  & L  & L  & L  & L  \\
1\textsc{pl.sbjv} & L  & NL$^*$ & L  & L  & L  & L  \\
2\textsc{pl.sbjv} & L  & NL$^*$ & L  & L  & L  & L  \\
3\textsc{pl.sbjv} & L  & L  & L  & L  & L  & L  \\
\hline
\end{tabular}
}
\caption{$K$-means cluster assignments in the 10\%L condition. Column abbreviations: Van = \textsc{Vanilla}, C-Sep = \textsc{Character-separated}, F-Inv = \textsc{Feature-invariant}, F-1H = \textsc{Feature-onehot}, F-Geo = \textsc{Feature-geometric}, Exp.\ = expected cluster under a perfect L-shape. L = L-shape cluster, NL = non-L cluster. $^*$Misclassified cells.}
\label{tab:kmeans-10L}
\end{table}

All three Class 2 models and \textsc{Character-separated} correctly cluster 1\textsc{sg.ind} with all six subjunctive cells. \textsc{Vanilla} achieves only a partial L-shape: 1\textsc{pl.sbjv} and 2\textsc{pl.sbjv} are misclassified with the non-L indicative cells. Although \textsc{Character-separated} achieves the correct grouping, its L-shape scores are lower than those of the Class 2 models, indicating weaker paradigm differentiation.  

\subsection{Nonce verb evaluation}

\subsubsection{Stem accuracies: models vs.\ humans}

\begin{figure}[ht]
\centering
  \includegraphics[width=\columnwidth]{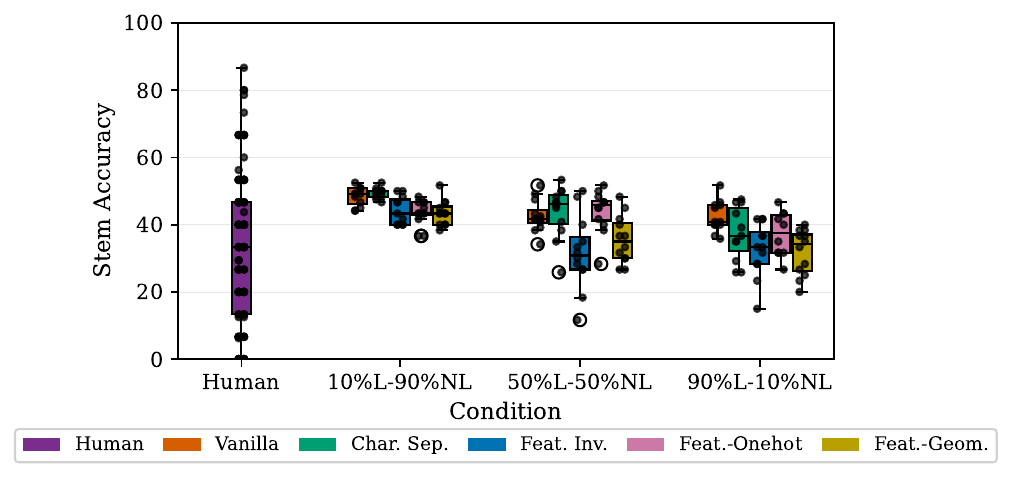}
  \caption{Stem accuracy on nonce verbs from \citet{Nevins2015TheRA}. Human results (left) are shown alongside model results across the three frequency conditions.}
  \label{fig:nonce-stem-accuracy}
\end{figure}

Figure~\ref{fig:nonce-stem-accuracy} compares relaxed stem accuracy on the 15 nonce verbs from \citet{Nevins2015TheRA} across all five architectures and human participants. All models outperform the human mean of 33\%: in the 10\%L condition, \textsc{Character-separated} and \textsc{Vanilla} reach 49\% and 48\% respectively, while the three Class 2 models cluster around 43-44\%.

Unlike the real-verb results, increasing L-shaped training frequency does not improve nonce-verb stem accuracy. All architectures decline from 10\%L to 90\%L, with Class 2 models dropping more sharply (from 43-44\% to 32-37\%) than Class 1 models (from 48-49\% to 37-42\%). This pattern, consistent with \citet{ramarao-etal-2025-frequency}, suggests that models trained on higher proportions of L-shaped verbs generalize less effectively to novel stems.

\subsubsection{Paradigm shape: models vs.\ humans}

\begin{figure}[ht]
\centering
      \includegraphics[width=\columnwidth]{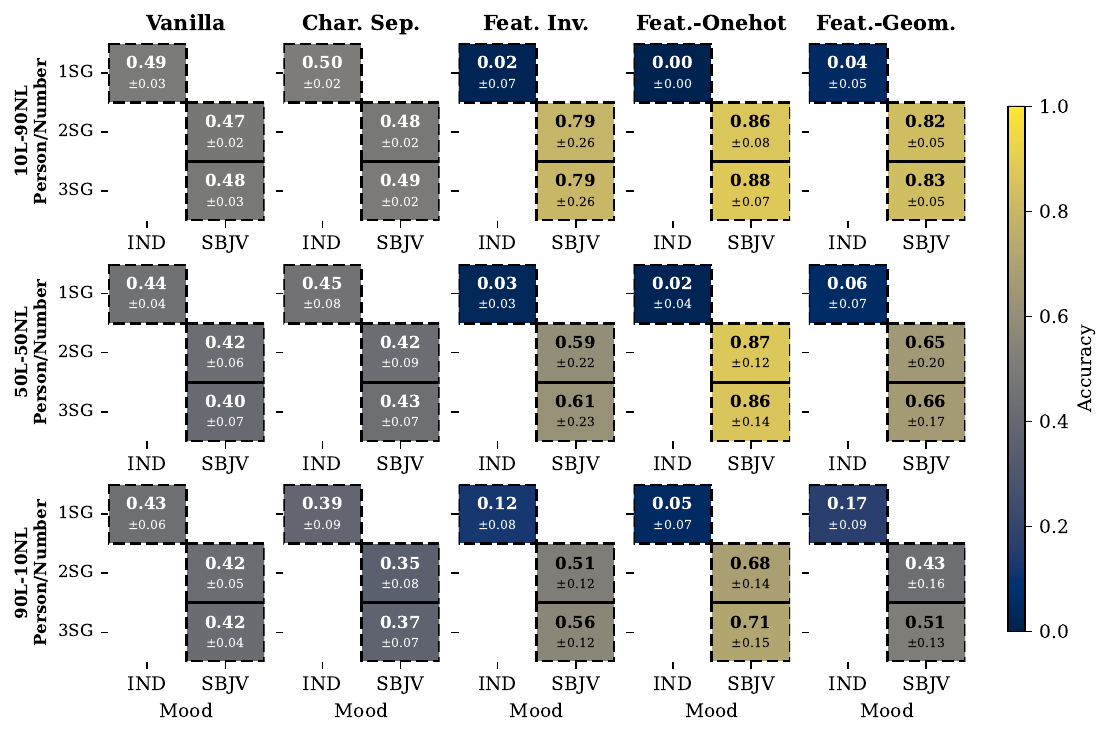}
  \caption{Per-cell mean stem accuracy on nonce verbs for all five architectures across the three frequency conditions.}
  \label{fig:nonce-paradigm-models}
\end{figure}

\begin{figure}[ht]
\centering
  \includegraphics[width=0.4\columnwidth]{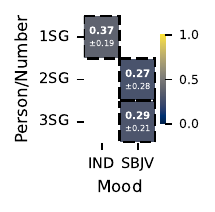}
  \caption{Per-cell mean stem accuracy on nonce verbs for human participants from \citet{Nevins2015TheRA}.}
  \label{fig:nonce-paradigm-humans}
\end{figure}

Figures \ref{fig:nonce-paradigm-models} and \ref{fig:nonce-paradigm-humans} show per-cell nonce-verb stem accuracy for models and humans, respectively. Class 1 models, Class 2 models, and humans each exhibit distinct paradigm shapes.
Human participants (Figure~\ref{fig:nonce-paradigm-humans}) show higher accuracy on 1\textsc{sg.ind} (37\%) than on subjunctive cells (27-29\%), suggesting that they preferentially generalize the stem alternation to 1\textsc{sg.ind}, the most frequent and salient L-shaped cell, rather than to subjunctive forms.

Class 2 models show the opposite pattern: subjunctive stem accuracy is high (79--88\% at 10\%L) while 1\textsc{sg.ind} accuracy is near zero (0-4\%). These models have learned to produce the L-shaped stem in subjunctive cells but fail to extend it to 1\textsc{sg.ind} for novel items, yielding a partial L-shape that captures the mood distinction but misses the defining feature of the morphome, the inclusion of 1\textsc{sg.ind}. Class 1 models (\textsc{Vanilla}, \textsc{Character-separated}) show near-uniform accuracy across all cells (47-50\% at 10\%L), with no preference for indicative or subjunctive. These models do not exhibit any paradigm shape on nonce verbs.

\section{Discussion}

\paragraph{Positional encoding and tag representation (RQ1).} Of the two architectural dimensions we manipulated, positional encoding proves decisive. When L-shaped verbs are scarce in training, position-invariant models recover higher L-shaped accuracy than sequential-PE models across all metrics. How tags are represented, whether as single indivisible vocabulary entries (e.g., \texttt{V;IND;PRS;1;SG}), one-hot feature vectors, or linguistically motivated feature geometries, matters far less: the three position-invariant architectures perform comparably despite substantial differences in how they encode morphosyntactic information. As L-shaped training frequency increases, all architectures converge, suggesting that position-invariant encoding acts as an inductive bias that compensates for data sparsity. Across all architectures, the minority verb class in training is consistently predicted poorly, confirming the frequency dependence shown by \citet{ramarao-etal-2025-frequency}. However, position-invariant models maintain higher L-shaped accuracy than sequential-PE models under the same low-frequency conditions, suggesting that the reliance on input frequency that \citet{ramarao-etal-2025-frequency} observed is a property of the vanilla architecture rather than of neural inflection models in general.

\paragraph{Structural pattern acquisition (RQ2).} Position-invariant models recover the correct L-shaped clustering, grouping 1\textsc{sg.ind} with all subjunctive cells, even from limited exposure, whereas sequential-PE models either fail to capture the pattern or do so only partially. The per-cell stem accuracy heatmaps confirm this: Class 2 models show a clear accuracy contrast between L-cells and the remaining indicative cells even at 10\%L, whereas Class 1 models show more uniform accuracy across cells. This suggests that position-invariant models acquire the L-shape as a structural paradigmatic pattern rather than merely memorizing individual stem alternations.

\paragraph{Productive generalization and comparison with humans (RQ3).} Yet none of the models extend the L-shaped pattern productively to novel stems. Position-invariant models generalize the L-shaped stem to subjunctive cells but fail almost entirely on 1\textsc{sg.ind}, while humans do the opposite \citep{Nevins2015TheRA}. This likely reflects a difference between the two test sets: the real verb test set shares its stem alternation patterns with training, whereas the nonce verbs introduce consonant alternations absent from Spanish. Models succeed when alternation patterns are familiar but fail when both stems and alternations are novel, consistent with prior evidence that neural inflection models struggle to match human wug-test behavior \citep{corkery-etal-2019-yet, weissweiler-etal-2023-counting}. Neither model class reproduces the human paradigm shape, indicating that human morphological generalization differs qualitatively from that of the transformer architectures tested here.

\section{Conclusion}

We compared five encoder-decoder transformers on their ability to acquire and generalize the Spanish L-shaped morphome. The critical architectural distinction is between the two classes of models, not between individual architectures within each class: models within Class 1 (sequential PE) and within Class 2 (position-invariant PE) behave similarly regardless of how morphosyntactic tags are represented. Position-invariant positional encoding enables learning the L-shaped paradigm even when L-shaped verbs are scarce. However, no architecture productively generalizes this pattern to novel stems, and all models diverge qualitatively from human generalization behavior. These findings contribute to the debate on the cognitive status of morphomic patterns: the L-shape is statistically learnable from distributional evidence, but the generalization mechanisms of character-aware transformers differ from those of human speakers. Future work should probe models' internal representations to determine whether the L-shape is encoded but not realized in production, extend the analysis to Italian and Portuguese where experimental data exists \citep{Nevins2015TheRA, cappellaro2024cognitive}, and explore training regimes that better support generalization of minority patterns.

\section{Limitations}

Whether our findings extend to other Romance languages or to the complete paradigm remains untested. The nonce verb stimuli from \citet{Nevins2015TheRA} introduce fricative-stop alternations absent from Spanish, making it difficult to separate failures of paradigm generalization from failures on novel phonological alternations. All five models share the same hyperparameters from \citet{ramarao-etal-2025-frequency}, which may not be optimal for each architecture individually. Finally, we compare only transformer variants. Non-neural models such as linear discriminative learning have shown competitive performance on cognitively plausible morphological generalization tasks \citep{jeong-etal-2023-linear} and may behave differently on the acquisition and generalization of morphomic patterns.

\section{Bibliographical References}

\bibliographystyle{lrec2026-natbib}
\bibliography{custom}

\end{document}